\PassOptionsToPackage{table}{xcolor}
\documentclass[sigconf]{acmart}
\setcopyright{none}
\renewcommand\footnotetextcopyrightpermission[1]{}
\makeatletter
\renewcommand\@affiliationfont{\footnotesize\normalfont}
\makeatother

\usepackage{amsmath,amssymb,booktabs,graphicx,multirow}
\usepackage{cleveref}

\newcommand{\bestcell}[1]{\cellcolor{green!24}\textbf{#1}\textsuperscript{1}}
\newcommand{\secondcell}[1]{\cellcolor{yellow!34}#1\textsuperscript{2}}
\newcommand{\thirdcell}[1]{\cellcolor{orange!22}#1\textsuperscript{3}}

\usepackage{float}
\usepackage[most]{tcolorbox}
\title{A General-Purpose VLM Can Teach an Astronomy Foundation Model to Better Recognize Galaxy Morphology}
\acmConference[Conference '27]{}{August 2027}{}

\author{Dichang Zhang}
\affiliation{%
  \department{Department of Computer Science}
  \institution{Stony Brook University}
  \city{Stony Brook}
  \state{New York}
  \country{USA}}
\email{diczhang@cs.stonybrook.edu}

\author{Jiaqi Deng}
\affiliation{%
  \department{School of Computer Science}
  \institution{University of Technology Sydney}
  \city{Sydney}
  \state{New South Wales}
  \country{Australia}}
\email{jiaqi.deng@student.uts.edu.au}

\author{Yixuan Shao}
\affiliation{%
  \department{Department of Physics and Astronomy}
  \institution{Stony Brook University}
  \city{Stony Brook}
  \state{New York}
  \country{USA}}
\email{yixuan.shao@stonybrook.edu}

\author{Yuanpeng Liu}
\affiliation{%
  \department{Department of Computer Science}
  \institution{Stony Brook University}
  \city{Stony Brook}
  \state{New York}
  \country{USA}}
\email{yuanpliu@cs.stonybrook.edu}

\author{Jiali Cui}
\affiliation{%
  \institution{Futurewei Technologies}
  \city{San Jose}
  \state{California}
  \country{USA}}
\email{jcui2@futurewei.com}

\author{Zhiqiang Lao}
\affiliation{%
  \institution{Futurewei Technologies}
  \city{San Jose}
  \state{California}
  \country{USA}}
\email{zlao@futurewei.com}

\author{Heather Yu}
\affiliation{%
  \institution{Futurewei Technologies}
  \city{San Jose}
  \state{California}
  \country{USA}}
\email{hyu@futurewei.com}

\author{Liang Peng}
\affiliation{%
  \institution{Futurewei Technologies}
  \city{San Jose}
  \state{California}
  \country{USA}}
\email{lpeng@futurewei.com}

\author{Simon Birrer}
\affiliation{%
  \department{Department of Physics and Astronomy}
  \institution{Stony Brook University}
  \city{Stony Brook}
  \state{New York}
  \country{USA}}
\email{simon.birrer@stonybrook.edu}

\author{Dimitris Samaras}
\affiliation{%
  \department{Department of Computer Science}
  \institution{Stony Brook University}
  \city{Stony Brook}
  \state{New York}
  \country{USA}}
\email{samaras@cs.stonybrook.edu}
\renewcommand{\shortauthors}{Zhang et al.}

\ccsdesc[500]{Applied computing~Astronomy}
\ccsdesc[500]{Computing methodologies~Object recognition}

\keywords{ Observational Cosmology, Astronomical Survey, Image Classification, Vision-Language Model, Knowledge Distillation, Scientific Foundation Model}

\begin{teaserfigure}
\centering
\includegraphics[width=\textwidth]{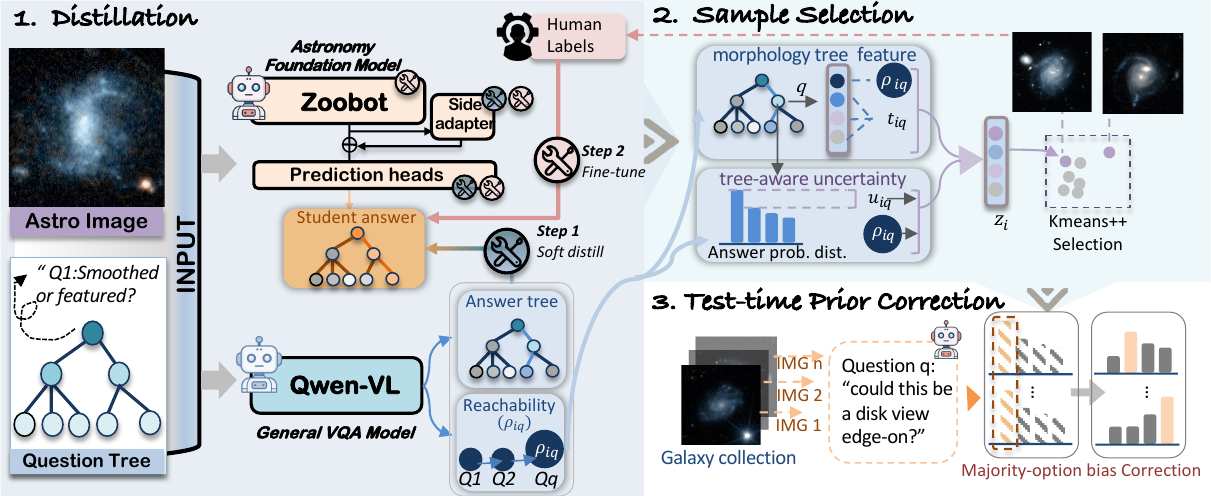}
\caption{
    Overview of the proposed VLM-teacher framework for label-efficient galaxy
morphology classification. A general-purpose VLM answers questions
over the morphology decision tree, and its soft predictions are
distilled into a Zoobot-based astronomy foundation model using
reachability-aware supervision and side tuning. The teacher's tree-probability
features and predictive uncertainty guide the selection of galaxies for human
annotation, after which the student is finetuned on the acquired labels.
Finally, test-time prior correction mitigates majority-option
bias.
}

\label{fig:teaser}
\Description{Overview of the VQA-teacher pipeline for survey morphology classification.}

\end{teaserfigure}

\begin{document}
\begin{abstract}
The upcoming decade of observational cosmology will be shaped by large sky
surveys that will deliver unprecedented volumes of galaxy imaging data.
Existing astronomy foundation models provide strong galaxy representations,
but adapting them to new survey conditions and survey-specific morphology recognition tasks still requires substantial human supervision. In this work, we show that
vision-language-model (VLM)-based visual question answering (VQA) systems
contain meaningful visual-semantic priors that can serve as weak supervision
for downstream morphology classifiers and improve morphology classification
under limited human-label budgets.
We first introduce a survey-oriented VQA benchmark spanning two
representative imaging regimes and evaluate state-of-the-art VLMs on galaxy
morphology questions. The results show that these models capture useful
morphology signals and informative uncertainty, but are not sufficiently
reliable to replace human annotators. Motivated by this finding, we use a
general-purpose VLM as a morphology teacher for Zoobot, an astronomy foundation
model pretrained on large-scale Galaxy Zoo annotations. The teacher's soft
predictions are distilled through a side-tuning module,
transferring visual-semantic knowledge while preserving Zoobot's pretrained
morphology representations. We further combine teacher uncertainty with
tree-probability features for uncertainty-and-diversity sample selection,
followed by human-label finetuning and test-time prior correction.
Across two survey domains and multiple annotation budgets, the VLM teacher
consistently improves Zoobot's downstream morphology classification. These results
demonstrate that a general-purpose VLM provides knowledge complementary to an
astronomy foundation model and can teach it to better recognize galaxy
morphology under limited human supervision. The resulting pipeline is designed
for label-efficient adaptation to forthcoming large-scale surveys, including the Vera C. Rubin Observatory’s Legacy Survey of Space and Time (LSST) and the Nancy Grace Roman Space Telescope. The benchmark and code are publicly available at
\url{https://github.com/fw-ic/VLM-morphology-teacher}.
\end{abstract}
    
\maketitle
\section{Introduction}
\label{sec:intro}
Observational cosmology is entering an era defined by large sky surveys. Unlike traditional pointed observations, which target individual objects or small fields, modern sky surveys systematically image large fractions of the sky and automatically catalog hundreds of millions to billions of sources. Ground-based facilities observe in the optical bands under atmospheric seeing.
The Vera C.~Rubin Observatory's Legacy Survey of Space and Time
(LSST; \cite{ivezic2019lsst}) builds on earlier deep-wide optical imaging
surveys such as the Dark Energy Survey (DES)~\cite{dey2019legacy} and the Hyper Suprime-Cam Survey (HSC)~\cite{Aihara2018HSC},
will provide deep, wide-field optical imaging over much of the southern sky. It will image billions of galaxies with a point-spread function (PSF) of $\sim$0.7--1.0 arcsec. Space-based missions operate above the atmosphere, achieving sharper PSFs (${\lesssim}0.2$ arcsec) and extending sensitivity into the near-infrared: ESA's Euclid \cite{laureijs2011euclid} and NASA's Nancy Grace Roman Space Telescope \cite{akeson2019roman} together plan to cover $\sim$15,000 deg$^2$. This survey regime creates two coupled challenges for morphology analysis. First, survey images are not uniformly high-resolution, high-signal-to-noise targeted observations: ground-based surveys are limited by atmospheric seeing, while all surveys must contend with variations in depth, noise, wavelength coverage, and PSF across instruments. Second, the number of galaxies increases by orders of magnitude, making object-by-object visual inspection infeasible.
Galaxy morphology---whether a galaxy is a smooth elliptical or a featured disk, and whether it contains bars, spiral arms, bulges, clumps, or merger signatures---is a key observable for studying galaxy formation and evolution and for measuring the demographics of galaxy populations across cosmic time. Both ground- and space-based surveys must therefore measure morphology at scale, across heterogeneous imaging conditions. Citizen-science efforts such as Galaxy Zoo \cite{lintott2008galaxyzoo,willett2013gz2} have produced rich morphological catalogs by asking volunteers to answer a branched decision tree of questions. However, exhaustive manual labeling cannot scale to the data volumes expected from next-generation surveys, and the morphology of most survey galaxies will never be inspected by a human.

\begin{figure}[h]
    \centering
    \includegraphics[width=\linewidth]{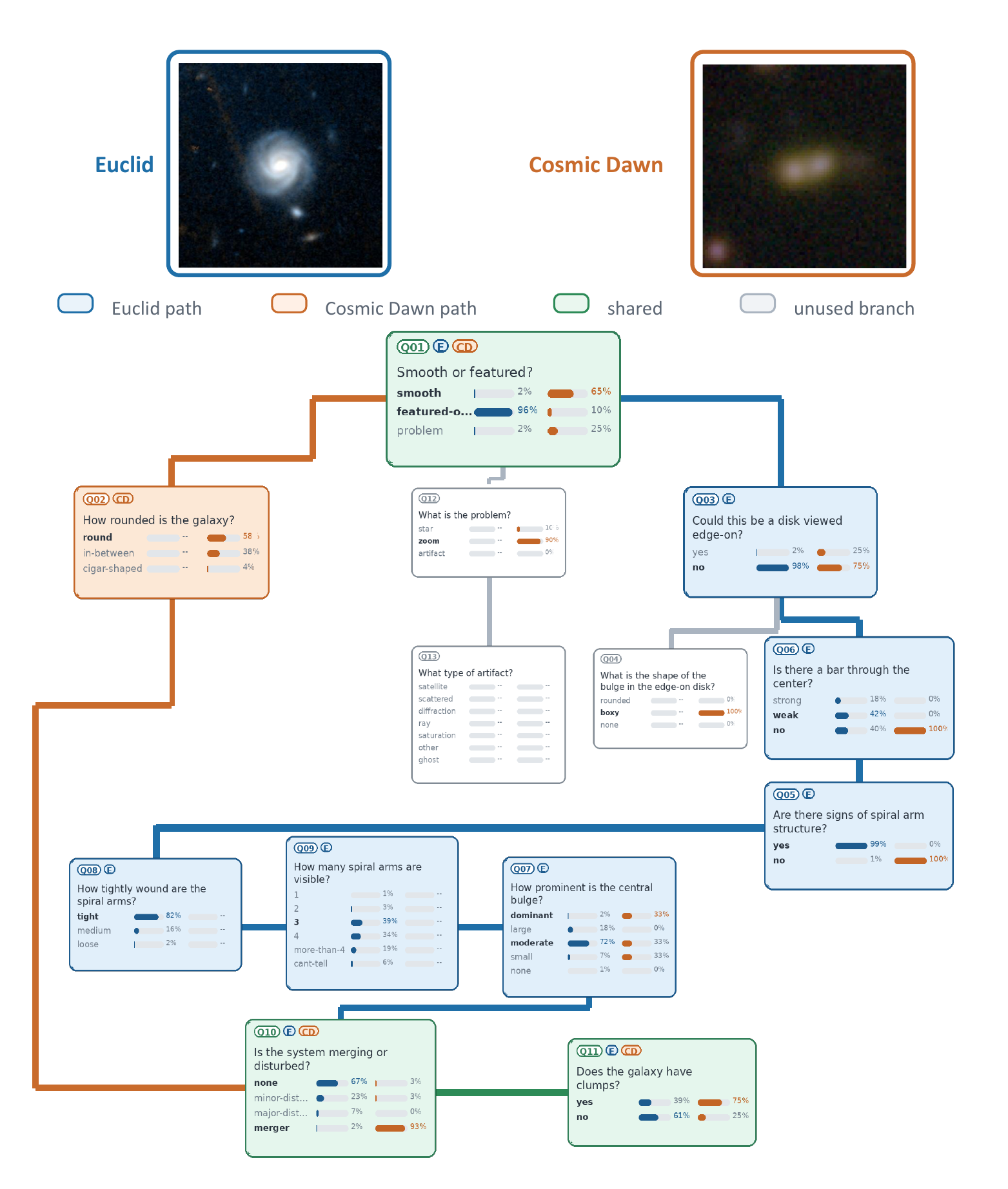}
    \caption{
    Euclid and Cosmic Dawn examples with their paths through a morphology question tree.
    Top: representative cutouts from Euclid and Cosmic Dawn.
    Bottom: selected branches in the question tree; blue traces the Euclid path, orange traces the Cosmic Dawn path, green marks shared downstream questions, and gray marks unused branches.
    Node bars show vote fractions for both examples.
    }
    \label{fig:data-example}
\end{figure}

The prevailing approach is to train image classifiers on large sets of human annotations and then apply them at survey scale~\cite{dieleman2015rotation,dominguezsanchez2018,walmsley2020bayesian,walmsley2022decals}. This line of work has led to general-purpose morphology models such as Zoobot~\cite{walmsley2023zoobot}, which are pretrained on large Galaxy Zoo~\cite{lintott2008galaxyzoo} annotation sets spanning multiple telescopes and can be adapted to new survey domains~\cite{huertascompany2023review}. 

Despite their effectiveness, two related limitations remain. First, morphology
models pretrained only on existing telescope  images cannot
cover the full range of observing conditions encountered in future surveys.
Differences in angular resolution, wavelength coverage, point-spread function,
depth, noise properties, and source populations create visual domain shifts
that require survey-specific adaptation. Second, Galaxy Zoo projects often
modify their morphology question trees to reflect the structures visible in
each survey and its scientific objectives. Models such as Zoobot, however,
learn mappings from images to the predefined outputs of a fixed annotation
tree rather than explicitly modeling the semantics of its questions. Changes
to the questions, answer choices, or branching structure therefore typically
require a new task-specific prediction head and additional human annotations. 
Adapting these models to a new survey and annotation protocol consequently
remains supervision-intensive.

Modern VLM-based VQA systems \cite{antol2015vqa,radford2021clip,liu2023llava} offer a complementary set of capabilities. Through large-scale image-text pretraining, they acquire broad visual and semantic knowledge and can answer diverse natural-language questions about images under varying visual conditions without task-specific training. This raises two natural questions. First, can general-purpose VLM-based VQA systems answer galaxy morphology questions accurately enough to serve as standalone classifiers? Second, if their zero-shot predictions are not sufficiently reliable, can they nevertheless provide useful teacher signals that can be distilled \cite{hinton2015distilling} into a domain-specific model such as Zoobot, helping it adapt beyond its pretrained survey and annotation settings while making more effective use of a limited human-label budget?

The underlying intuition is that many Galaxy Zoo questions depend partly on broadly learned visual concepts rather than exclusively on astronomy-specific expertise. A non-expert who has never studied survey imagery can still reason about whether an object appears round, elongated, symmetric, or irregular. Similarly, language-image pretraining can give semantic meaning to concepts such as a spiral arm, bar, clump, or merger before the model has observed many labeled galaxy examples. These visual-linguistic priors may complement the survey-specific representations learned by galaxy-image foundation models \cite{walmsley2023zoobot,parker2024astroclip}.

To test this idea, we proceed in two steps. First, we build a survey-oriented Galaxy morphology VQA benchmark over two representative future-survey-like datasets: Cosmic Dawn~\cite{pearson2026cosmicdawn}, which approximates LSST-like ground-based survey imaging, and Euclid Q1~\cite{euclid2025q1morphology}, which represents space-based survey imaging with optical and near-infrared information. Detail comparison between Cosmic Dawn, Euclid, LSST, and Roman are provided in Appendix~\ref{app:dataset-build} \cref{tab:survey-comparison}. Representative examples from the two datasets, together with their corresponding paths through the morphology question tree, are shown in \cref{fig:data-example}. We evaluate three state-of-the-art open VLM-based VQA systems in a zero-shot setting and measure accuracy and confidence calibration. These results show that current VQA models are not reliable enough to serve as stand-alone morphology classifiers, but they do have meaningful parametric knowledge for morphology classification: across most questions, they perform substantially above random-choice baselines. We further find that their confidence scores are reasonably calibrated, suggesting that the models often know when they are uncertain. The benchmark therefore identifies general-purpose VQA models as useful but insufficient sources of morphology supervision.

Second, motivated by this finding, we develop a two-stage teacher-student framework that uses a general-purpose VQA model as a morphology teacher for a domain-specific classifier, instantiated with Zoobot. In Stage~1, we distill the teacher's soft per-question predictions on a large unlabeled image pool into a side-tuning module, allowing the student to acquire the teacher's visual-semantic priors while preserving its pretrained morphology representations. In Stage~2, the student is finetuned using the available human annotations. We further propose a tree-aware sample-selection strategy that combines the teacher's predictive uncertainty with tree-probability features, yielding an uncertainty-and-diversity criterion for selecting the most informative galaxies for human annotation. Finally, at inference time, we apply a lightweight prior-correction module that compensates for the majority-option bias of the distilled classifier, improving performance on underrepresented morphology classes.

Across varying human-label budgets, the VQA teacher consistently improves the performance of the Zoobot-based classifier. In summary, our contributions are:
(1) a survey-oriented Galaxy Zoo VQA benchmark spanning two representative survey regimes, together with an empirical study of state-of-the-art VQA systems as galaxy morphology answerers, characterizing their strengths, failure modes, and uncertainty behavior;
(2) a VQA-teacher framework that combines side-tuning-based soft-label distillation, tree-aware uncertainty-and-diversity sample selection, and test-time prior correction to improve the label efficiency of galaxy morphology classification;
(3) an openly released, deployment-oriented package comprising VQA-ready, post-processed Galaxy Zoo benchmarks, precomputed logits from multiple open VLMs, and a reproducible implementation of our full pipeline. The package enables researchers to use, study, reproduce, and extend VQA-assisted galaxy morphology methods without repeating costly large-model inference, while providing an end-to-end, label-efficient pipeline that can be readily adapted and deployed for forthcoming large-scale surveys such as LSST and Roman.

\section{Related work}

\subsection{Galaxy morphology classification}
Galaxy morphology is central to galaxy demographics, where survey-scale
classification enables population-level measurements of the prevalence and
diversity of morphological structures. Galaxy Zoo  established crowd-sourced visual
morphology classification at scale by asking volunteers to annotate galaxy
images. It began with relatively coarse classifications of Sloan Digital Sky
Survey images~\cite{lintott2008galaxyzoo} and later adopted a branched decision
tree of morphology questions, with volunteer responses aggregated into
per-question vote fractions~\cite{willett2013gz2}. Subsequent Galaxy Zoo
campaigns extended this paradigm to deeper and higher-resolution imaging,
including Hubble~\cite{willett2017gzhubble}, CANDELS~\cite{simmons2017gzcandels}, DECaLS~\cite{walmsley2022decals}, Cosmic
Dawn~\cite{pearson2026cosmicdawn}, and Euclid Q1 visual morphology
catalogue~\cite{euclid2025q1morphology}. These efforts demonstrate both the
scientific value of detailed visual morphology and the growing cost of
obtaining it as surveys increase in depth, area, and wavelength coverage.

Importantly, these projects do not share an identical annotation protocol.
Their question trees are revised according to the structures visible in each
dataset and the scientific goals of the survey. For example, Galaxy Zoo 2
expanded the Galaxy Zoo 1 labels into a branched tree covering bars,
bulge prominence, spiral-arm winding and number, and edge-on disk structure
\cite{lintott2008galaxyzoo,willett2013gz2}. Galaxy Zoo: CANDELS
introduced questions targeting clumpiness, bar instabilities, and merger or
tidal signatures that are especially relevant in deeper, higher-redshift
Hubble imaging \cite{simmons2017gzcandels}. Galaxy Zoo DECaLS also revised
answer sets and branches relative to Galaxy Zoo 2, including changes to the
bulge, spiral-arm, and interaction-related tasks
\cite{walmsley2022decals}. Thus, transfer across Galaxy Zoo datasets involves
not only visual domain shift but also changes in question semantics, answer
spaces, and tree structure. 


\subsection{Astronomical foundation models}
Self-supervised learning has shown that large collections of unlabeled
telescope images can provide transferable representations for downstream
astronomical tasks~\cite{hayat2021selfsupervised}. Multimodal astronomical
foundation models extend this idea by combining complementary observational
modalities. For example, AstroCLIP~\cite{parker2024astroclip} aligns galaxy
images and spectra in a shared representation space and supports downstream
tasks including morphology classification.

Large and diverse Galaxy Zoo annotation sets provide another route to reusable
galaxy representations. Zoobot~\cite{walmsley2023zoobot} is pretrained on
morphology annotations collected across multiple telescopes and question trees,
and can subsequently be adapted to new survey domains. This supervised
pretraining produces strong morphology-specific visual representations, but
its knowledge is still derived from a fixed collection of astronomical images
and annotation protocols.

Our work is complementary to these galaxy-specific foundation models. Rather
than learning only from astronomical images, spectra, or existing Galaxy Zoo
annotations, we investigate whether a general-purpose vision-language model,
trained largely on data outside astronomy, contains visual-semantic priors that
can improve galaxy morphology recognition. We use Zoobot as the
astronomy-specific student and distill into it weak supervision from a
general-purpose VQA teacher. The resulting framework combines Zoobot's
domain-specific morphology representations with the broader visual and
linguistic knowledge of the VQA teacher.

\subsection{Vision-language models and visual question answering}

Recent advances in multimodal learning have led to the development of vision-language models (VLMs), which jointly process visual and textual information~\cite{radford2021clip,alayrac2022flamingo,li2023blip2,liu2023llava}. They have substantially improved cross-modal understanding and enabled strong zero-shot transfer across diverse vision-language tasks. Among which, visual question answering (VQA) is one of the most standard measurer for evaluating VLM's fundamental abilities~\cite{antol2015vqa}. As VLM capabilities continue to advance, large-scale benchmarks such as ScienceQA~\cite{lu2022scienceqa} and MMMU~\cite{yue2024mmmu} have been further proposed to assess domain-specific and expert-level multimodal reasoning beyond understanding commonsense visual concepts. In this spirit, we benchmark representative open-source VLM to evaluate their performance on our domain-specific VQA tasks.
\section{Morphology VQA benchmark}
\label{sec:benchmark}

\subsection{Survey-oriented datasets}

We construct a survey-oriented VQA benchmark for evaluating general-purpose
VLMs on structured galaxy morphology questions. The benchmark covers two
representative imaging regimes. Cosmic Dawn~\cite{pearson2026cosmicdawn}
provides deep ground-based optical imaging and approximates conditions expected
in future LSST observations, while Euclid Q1~\cite{euclid2025q1morphology}
provides high-resolution space-based optical/near-infrared imaging relevant to
future surveys such as Roman.

Each dataset contains approximately $45{,}000$ Galaxy-Zoo-labeled galaxies
represented by $224$ px image cutouts. For each galaxy, we evaluate only the
questions reachable along its reference morphology-tree path.
\Cref{fig:data-example} shows representative examples from both surveys and
their corresponding paths through a morphology question tree. Details of dataset construction, together with a detailed comparison between the benchmark
surveys and forthcoming flagship surveys, are provided in
Appendix~\ref{app:dataset-build}.

The natural catalogue distribution is dominated by common root answers and
does not reliably expose failures on rare downstream branches. We therefore
construct a diagnostic test set that balances reachable question--answer
strata. The resulting set contains
1,066 Cosmic Dawn and 1,168 Euclid galaxies and covers 44 and 43 reachable
strata, respectively. The detailed greedy construction procedure is provided
in Appendix~\ref{app:tree-balanced-test}.

\subsection{VQA evaluation}
\label{sec:vqa-evaluation}

We evaluate three open VQA models as galaxy-morphology answerers:
Qwen3.5-9B~\cite{wang2024qwen2vl}, InternVL3.5-8B
Flash~\cite{chen2024internvl}, and Llama3.2-11B
Vision~\cite{grattafiori2024llama3}. For each galaxy, the model is queried question-by-question along the decision tree. At each reachable node, it receives the image, the natural-language question, the legal answer options, and the oracle path history, and produces an option-level confidence distribution. We evaluate both hard-answer accuracy and agreement between the predicted distribution and the source vote-fraction vector.

We report three accuracy metrics. \textbf{Overall} accuracy pools all evaluated
question instances. \textbf{Balanced} accuracy macro-averages, over the 13
questions, option-balanced accuracy in which every observed option contributes
equally regardless of frequency. \textbf{Weak} accuracy averages accuracy over
four difficult downstream questions: artifact type, bulge size, spiral
winding, and spiral arm count.
The full VLM evaluation protocol and mathematical definitions of these accuracy
and distribution metrics are provided in Appendix~\ref{app:vqa-eval-metrics}.

\Cref{tab:zero-shot-summary} summarizes all-method performance.
Qwen is the strongest VLM across all reported metrics, reaching 52.19\% and
57.36\% overall accuracy on Cosmic Dawn and Euclid, respectively. These results
show that general-purpose VQA systems contain nontrivial morphology knowledge
despite not being trained specifically for this task. However, the models are not
reliable enough to serve as stand-alone morphology annotators.

\begin{table}[t]
  \centering
  \scriptsize
  \caption{Zero-shot VLM accuracy. Random is the uniform answer-choice baseline.}
  \label{tab:zero-shot-summary}
  \begin{tabular}{llrrr}
    \toprule
    Survey & Model & Overall (\%) $\uparrow$ & Balanced (\%) $\uparrow$ & Weak (\%) $\uparrow$ \\
    \midrule
    \multirow{5}{*}{Cosmic Dawn}
      & Qwen3.5-9B             & 52.19 & 42.51 & 29.82 \\
      & InternVL3.5-8B Flash   & 46.90 & 38.89 & 22.45 \\
      & Llama3.2-11B Vision    & 33.64 & 35.87 & 16.59 \\
      \cmidrule(lr){2-5}
      & Random                 & 34.95 & 32.77 & 21.07 \\
      & Human                  & 73.17 & 72.64 & 63.39 \\
    \midrule
    \multirow{5}{*}{Euclid}
      & Qwen3.5-9B             & 57.36 & 47.57 & 29.31 \\
      & InternVL3.5-8B Flash   & 52.30 & 44.56 & 23.86 \\
      & Llama3.2-11B Vision    & 40.23 & 41.61 & 25.81 \\
      \cmidrule(lr){2-5}
      & Random                 & 33.71 & 32.77 & 21.07 \\
      & Human                  & 68.96 & 63.46 & 54.13 \\
    \bottomrule
  \end{tabular}
\end{table}

We additionally examine how model performance varies with scale. \Cref{tab:model-scaling} reports the performance changes of different Qwen3.5 model sizes relative to the 9B dense model. Overall, increasing model size does not consistently improve performance on our task.

\begin{table}[t]
  \centering
  \caption{Scaling ablation. Values are signed percentage-point changes relative to 9B dense model, averaged equally over Cosmic Dawn and Euclid}
  \label{tab:model-scaling}
  \small
  \begin{tabular}{lrrr}
    \toprule
    Model & $\Delta$Overall (\%) $\uparrow$ & $\Delta$Balanced (\%) $\uparrow$ & $\Delta$Weak (\%) $\uparrow$ \\
    \midrule
    27B (dense)      & $-1.09$ & $-0.17$ & $-1.44$ \\
    35B-A3B (MoE)    & $-1.30$ & $-0.64$ & $+2.90$ \\
    122B-A10B (MoE)  & $+0.59$ & $-0.86$ & $-1.44$ \\
    \bottomrule
  \end{tabular}
\end{table}



\begin{table}[t]
  \centering
  \caption{Prompt ablation. Values are signed percentage-point changes relative to V0 Basic, averaged equally over Cosmic Dawn and Euclid.}
  \label{tab:prompt-variants}
  \scriptsize
  \begin{tabular}{lrrr}
    \toprule
    Variant & $\Delta$Overall (\%) $\uparrow$ & $\Delta$Balanced (\%) $\uparrow$ & $\Delta$Weak (\%) $\uparrow$ \\
    \midrule
    V1 Phenomenology      & $-0.74$ & $+0.82$ & $+0.27$ \\
    V2 Pheno. + Physics   & $-0.53$ & $+0.21$ & $+2.06$ \\
    V3 Evidence First     & $-4.24$ & $-1.19$ & $-0.86$ \\
    V4 Pheno. + Physics + Context       & $-0.45$ & $+0.49$ & $+3.01$ \\
    \bottomrule
  \end{tabular}
\end{table}

We also use a basic prompt as the reference and evaluate four variants that add
visible-morphology descriptions, physical grounding, an evidence-first
reasoning step, or survey-specific context. The prompt structures are generic
templates, while the morphology- and physics-specific content is supplied by our
domain experts. Full prompt templates are provided in
Appendix~\ref{app:prompt-designs}. Their survey-, question-, and
option-specific placeholders are completed by our domain experts, and the fully
instantiated executable prompts are included in our open-source codebase.

\Cref{tab:prompt-variants} reveals a trade-off between population-weighted
performance and performance on underrepresented morphology branches. The basic
prompt (V0) achieves the strongest overall accuracy, while V1 Phenomenology
slightly sacrifices overall performance in exchange for gains in balanced and
weak-branch accuracy. Adding survey-specific context in V4 further strengthens
this effect, producing the largest improvement on weak branches. These results suggest that V0, V1, and V4 provide different
operating points depending on whether the priority is aggregate accuracy or
performance on rare and difficult morphology categories.

\begin{figure}[h]
    \centering
    \includegraphics[width=\linewidth]{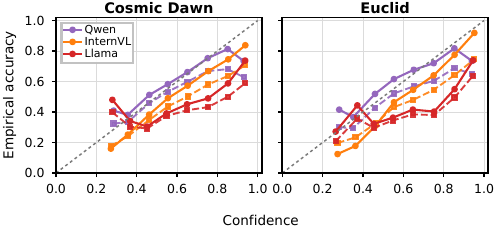}
   \caption{
Confidence–performance relations.
Each point shows one confidence bin, with the $x$-coordinate giving the predicted confidence and the $y$-coordinate giving empirical accuracy. Solid lines show agreements with hard accuracy, while dashed lines show agreements with  human vote fractions.
}
\label{fig:confidence-v-acc}
\end{figure}

We next evaluate whether the VQA models' logit-derived choice confidence is informative. To avoid relying on free-form verbal confidence estimates, we cast each Galaxy Zoo question as a multiple-choice VQA prompt: the valid answer options are mapped to single-letter choices (A, B, C, \ldots), and the model is instructed to output only the corresponding letter. We then read the logits assigned to the valid letter tokens at the answer position and apply a softmax over these logits to obtain an option-level confidence distribution. The predicted confidence is the probability assigned to the selected option. \Cref{fig:confidence-v-acc} plots empirical accuracy against this predicted confidence on the reference decision-tree path, with separate curves for Cosmic Dawn and Euclid. The curves are generally monotonic: higher logit-derived confidence corresponds to higher empirical accuracy for most models and both surveys. The dashed curves, which compare predicted confidence with agreement under the source fraction vector, follow a similar trend.

Taken together, the benchmark supports two conclusions. First, zero-shot VQA systems contain useful
morphology knowledge of both ground- and space-based survey regimes, but their raw predictions
are not sufficiently reliable for direct catalogue construction. Second, their confidence estimates
are informative enough to support downstream use. These findings
motivate the next stage of our method: distilling a VQA teacher into a dedicated morphology
classifier and using the teacher's uncertainty to guide which galaxies should receive scarce human
labels.

\section{Method}
\label{sec:optwt}

\subsection{Distillation}
\label{sec:distillation}

We use Qwen as the teacher and a pretrained Zoobot encoder as the student.
For each image $x_i$ and question $q\in\mathcal{Q}$ with answer set
$\mathcal{C}_q$, the teacher produces a soft probability vector
$t_{iq}\in\Delta^{|\mathcal{C}_q|-1}$. The student predicts
\begin{equation}
s_{\theta,q}(x_i)
=
\operatorname{softmax}\!\left(g_{\theta,q}(x_i)\right),
\qquad q\in\mathcal{Q},
\end{equation}
where $g_{\theta,q}$ is the question-specific head applied to the Zoobot
representation.

Because the morphology questions form a conditional tree, not every question is
applicable to every galaxy. We therefore estimate question reachability from
the teacher probabilities, avoiding hard routing through uncertain upstream
predictions. The calibration analysis in \cref{fig:confidence-v-acc} shows that
the teacher confidence is informative, supporting its use for soft routing and
distillation.

For a non-root question $q$, let
$\mathcal{P}(q)=\{(r,c)\}$ denote the ancestor-question and required-answer
pairs along the path from the root to $q$. We define the teacher reachability
weight
\begin{equation}
\rho_{iq}
=
\prod_{(r,c)\in\mathcal{P}(q)} t_{ir,c},
\qquad
\rho_{i,\mathrm{root}}=1.
\label{eq:reachability}
\end{equation}
For example, \emph{how-rounded} is reachable in proportion to the teacher
probability of \emph{smooth}, while questions about bars, bulges, and spiral
arms additionally depend on the relevant upstream disk and orientation
decisions. Thus, downstream questions contribute according to the probability
that their parent path is active.

Let $\mathcal{U}$ denote the unlabeled training pool. We distill the teacher's
soft predictions using question-normalized, reachability-weighted cross
entropy:
\begin{equation}
\mathcal{L}_{\mathrm{distill}}(\theta)
=
\frac{1}{|\mathcal{Q}|}
\sum_{q\in\mathcal{Q}}
\frac{
\sum_{i\in\mathcal{U}}
\rho_{iq}\,
\operatorname{CE}\!\left(t_{iq},s_{\theta,q}(x_i)\right)
}{
\sum_{i\in\mathcal{U}}\rho_{iq}+\epsilon
},
\label{eq:distill}
\end{equation}
where
\begin{equation}
\operatorname{CE}(t_{iq},s_{iq})
=
-\sum_{c\in\mathcal{C}_q}
t_{iq,c}\log s_{iq,c}.
\end{equation}
The per-question normalization prevents frequently reached shallow questions
from dominating the objective.

\subsection{Residual side tuning}
\label{sec:side-network}

To preserve Zoobot's domain specific knowledge during distillation, we adopt a
feature-level variant of side tuning~\cite{zhang2020sidetuning}: the pretrained
encoder is frozen, while a trainable residual branch absorbs the VLM-derived
update. Let $E_0$ denote the pretrained Zoobot encoder and
$h_i=E_0(x_i)\in\mathbb{R}^d$. We define a bottleneck side adapter
\begin{equation}
A_{\phi}(h)
=
W_2\operatorname{GELU}\!\left(
W_1\operatorname{LayerNorm}(h)+b_1
\right)+b_2,
\qquad
\tilde h_i=h_i+A_{\phi}(h_i),
\label{eq:side-network}
\end{equation}

We initialize $(W_2,b_2)$ to zero, so that
$\tilde h_i=h_i$ at initialization, and attach an independent linear prediction
head to $\tilde h_i$ for each morphology question.

During the reachability-weighted distillation in \cref{eq:distill}, $E_0$
remains frozen and only the side adapter and question-specific heads are
optimized. The VLM-derived update is therefore learned through the residual
branch without modifying the pretrained encoder.

\subsection{Sample selection}
\label{sec:sample-selection}

After distillation, we use the VLM teacher to select a budget of $B$ galaxies
from a pool $\mathcal U$ for human annotation. 

Our goal is to select galaxies that are both uncertain under the teacher and
diverse across likely Galaxy Zoo morphology branches. 
Inspired by BADGE’s uncertainty-weighted diversity embeddings~\cite{ash2020badge}, We define a tree-probability feature that represents the teacher's predicted morphology state over the full tree and scale this feature by the teacher's image-level tree-aware uncertainty.

The tree-probability feature is defined as:
\begin{equation} \phi_i = \text{concat}_{q\in\mathcal{Q}}\left[ \rho_{iq}t_{iq} \right] \in \mathbb{R}^{\sum_{q\in\mathcal{Q}}|\mathcal{C}_q|} \label{eq:tree-probability-feature} \end{equation}
Here, $t_{iq}\in\Delta^{|\mathcal C_q|-1}$ is Qwen's predicted answer
distribution for question $q$, and $\rho_{iq}$ is Qwen's estimated probability
that question $q$ is reachable for galaxy $x_i$. The concatenation is taken in
a fixed question-tree order.  The resulting feature encodes
both the teacher's soft morphology prediction and the probabilistic structure
of the Galaxy Zoo question tree.

We next define a scalar teacher uncertainty for each galaxy. For each galaxy
and question, we use margin uncertainty:
\begin{equation} u_{iq}=1-\left(\operatorname{top}_1(t_{iq})-\operatorname{top}_2(t_{iq})\right), \label{eq:margin-uncertainty} \end{equation}
where $\operatorname{top}_1$ and $\operatorname{top}_2$ denote the largest and
second-largest probabilities in $t_{iq}$. We aggregate question-level
uncertainty using teacher reachability:
\begin{equation}
a_i=
\sum_{q\in\mathcal{Q}}
\rho_{iq} u_{iq}.
\label{eq:tree-margin}
\end{equation}

Finally, we combine the
tree-probability feature with the tree-aware uncertainty:
\begin{equation} z_i=\sqrt{a_i+\epsilon}\phi_i \in\mathbb{R}^{\sum_{q\in\mathcal{Q}}|\mathcal{C}_q|}. \label{eq:selection-embedding} \end{equation}
The uncertainty factor moves
high-uncertainty galaxies farther from the origin, while the concatenated
tree-probability feature spreads selected examples across different likely
morphology branches. We then select a batch of $B$ galaxies using k-means++~\cite{arthur2007kmeanspp}
seeding in this embedding space:
\begin{equation}
\mathcal S_B
=
\operatorname{kmeans++}_B
\left(
{z_i: x_i\in\mathcal U}
\right).
\label{eq:kmeans-selection}
\end{equation}

\begin{figure}[t]
    \centering
    \includegraphics[width=\columnwidth]{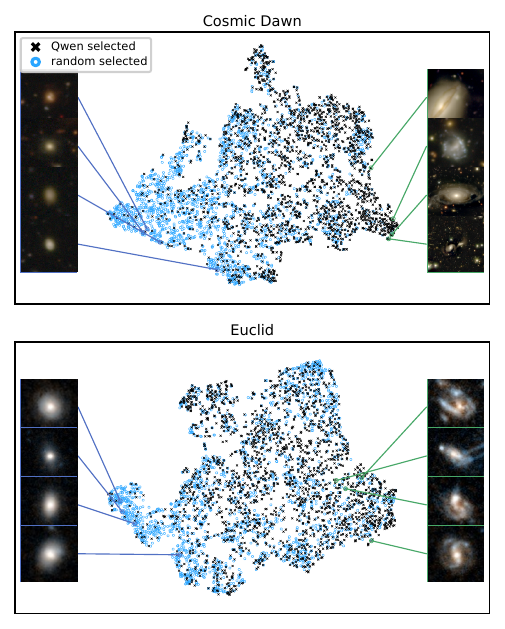}
    \vspace{-2mm}
    \caption{
    Visualization of the VLM selection space for Cosmic Dawn and Euclid at budget $B=2048$.
    Each panel shows a t-SNE embedding of the VLM selector feature.
    Black crosses indicate VLM-selected samples, and blue circles indicate randomly selected samples. Image callouts show representative galaxies from regions that are sparsely and densely sampled by the VLM selector. Compared with random sampling, VLM-guided selection allocates more samples to regions containing morphologically diverse and complex galaxies.
    }
    \label{fig:tsne-selection}
\end{figure}

\Cref{fig:tsne-selection} provides a qualitative visualization of the resulting teacher-guided acquisition behavior. We project the high-dimensional selection embeddings $z_i$ to two dimensions with t-SNE and compare the samples selected by the Qwen-guided strategy with a random subset of the same budget. In both Cosmic Dawn and Euclid, the Qwen-guided selector allocates more samples to regions corresponding to complex or morphologically diverse galaxies.

For the selected galaxies $i\in\mathcal S_B$, we obtain human Galaxy Zoo vote
fractions on the questions actually reached in the human annotation tree. Let
$h_{iq}$ denote the human vote-fraction target for question $q$, and let
$m_{iq}\in{0,1}$ indicate whether this human target is available for galaxy
$i$. We finetune the student with a question-balanced soft-label objective:
\begin{equation} \begin{aligned} \mathcal{L}_{\mathrm{ft}}(\theta) =& \sum_{q\in\mathcal{Q}} \frac{\sum_{i\in\mathcal{S}_B}m_{iq} \operatorname{CE}(h_{iq},s_{\theta,q}(x_i))} {\sum_{i\in\mathcal{S}_B}m_{iq}+\epsilon} \end{aligned} \label{eq:finetune} \end{equation}

\subsection{Test-time prior correction}

At evaluation time, we decode each Galaxy Zoo question independently. A plain argmax readout can favor frequent answers, because the model may assign high average probability to common options across the evaluation split. This is undesirable for identifying rare answers. We therefore apply a test-time prior correction to the student probabilities before taking the final answer.

For question $q$, let $\mathcal{E}_q$ be the set of evaluation examples on which question q is scored. Let $C_q$ be the answer set of question q. For each option $c \in C_q$, we first estimate the model-implied marginal option probability using the student predictions:
\begin{equation}
\tilde\pi_{q, c} = max(10^{-3}, |\mathcal{E}_q|^{-1} \sum_{i\in\mathcal{E}_q}s_{\theta,q,c}(x_i))    
\end{equation}

We then normalize these values within the option set of question q:
\begin{equation}
\hat\pi_{q, c} = \frac{\tilde\pi_{q, c}}{\sum_{c^\prime\in C_q}\tilde\pi_{q,c^\prime}}    
\end{equation}

Here, $\hat\pi_{q, c}$ is the model's own estimated answer prior on the evaluation examples for question q. The small floor 1e-3 prevents extremely small estimated priors from producing unstable corrections.

The final corrected readout is:
\begin{equation}
\hat{y}_{iq}^{(\beta)}
=
\arg\max_{c\in\mathcal{C}_q}
\frac{
s_{\theta,q,c}(x_i)
}{
\hat{\pi}_{q,c}^{\beta}
}.
\label{eq:beta-readout}
\end{equation}

The parameter $\beta$ controls the strength of the correction. When $\beta = 0$, the rule reduces to the ordinary argmax over the student probabilities. Larger beta values penalize answers that the model predicts too frequently across $E_q$, and therefore give relatively more weight to answers with smaller model-implied priors.

\section{Experiments}
\label{sec:results}
\begin{table}[t]
\centering\scriptsize
\setlength{\tabcolsep}{1.8pt}
\caption{Main comparison accuracy metrics. Finetune VLM is a single reported value. Colors mark rank by the mean within each survey-budget-metric block.}
\label{tab:main-results}
\resizebox{\columnwidth}{!}{%
\begin{tabular}{l r l c c c}
\toprule
Survey & Budget & Method & Overall (\%) \(\uparrow\) & Balanced (\%) \(\uparrow\) & Weak (\%) \(\uparrow\) \\
\midrule
Cosmic Dawn & 512 & No distil + random selection & 60.30{\tiny\,$\pm$\,1.12} & 52.51{\tiny\,$\pm$\,3.43} & 38.98{\tiny\,$\pm$\,3.69} \\
 &  & Finetune VLM + VLM selection & 55.57{\tiny\,$\pm$\,\raisebox{0.45ex}{\rule{2.10em}{0.28pt}}} & 47.72{\tiny\,$\pm$\,\raisebox{0.45ex}{\rule{2.10em}{0.28pt}}} & 29.84{\tiny\,$\pm$\,\raisebox{0.45ex}{\rule{2.10em}{0.28pt}}} \\
 &  & Naive distill + VLM selection & \secondcell{63.18{\tiny\,$\pm$\,0.13}} & \secondcell{54.32{\tiny\,$\pm$\,0.53}} & \thirdcell{39.03{\tiny\,$\pm$\,3.39}} \\
 &  & Ours + random selection & \thirdcell{61.85{\tiny\,$\pm$\,0.57}} & \thirdcell{52.98{\tiny\,$\pm$\,0.45}} & \secondcell{39.84{\tiny\,$\pm$\,3.92}} \\
 &  & Ours + VLM selection & \bestcell{66.31{\tiny\,$\pm$\,1.10}} & \bestcell{57.65{\tiny\,$\pm$\,0.59}} & \bestcell{44.45{\tiny\,$\pm$\,1.00}} \\
\addlinespace[1.0pt]
 & 2048 & No distil + random selection & 65.21{\tiny\,$\pm$\,1.01} & 56.50{\tiny\,$\pm$\,1.42} & \thirdcell{44.29{\tiny\,$\pm$\,5.10}} \\
 &  & Finetune VLM + VLM selection & 55.44{\tiny\,$\pm$\,\raisebox{0.45ex}{\rule{2.10em}{0.28pt}}} & 45.98{\tiny\,$\pm$\,\raisebox{0.45ex}{\rule{2.10em}{0.28pt}}} & 34.92{\tiny\,$\pm$\,\raisebox{0.45ex}{\rule{2.10em}{0.28pt}}} \\
 &  & Naive distill + VLM selection & \thirdcell{66.75{\tiny\,$\pm$\,0.58}} & \thirdcell{57.21{\tiny\,$\pm$\,0.88}} & 43.29{\tiny\,$\pm$\,1.88} \\
 &  & Ours + random selection & \secondcell{67.49{\tiny\,$\pm$\,0.63}} & \secondcell{58.59{\tiny\,$\pm$\,0.69}} & \secondcell{44.83{\tiny\,$\pm$\,1.51}} \\
 &  & Ours + VLM selection & \bestcell{69.97{\tiny\,$\pm$\,0.62}} & \bestcell{61.34{\tiny\,$\pm$\,0.11}} & \bestcell{47.86{\tiny\,$\pm$\,0.93}} \\
\addlinespace[1.0pt]
 & 8192 & No distil + random selection & 66.48{\tiny\,$\pm$\,1.05} & 56.97{\tiny\,$\pm$\,1.23} & 43.96{\tiny\,$\pm$\,4.67} \\
 &  & Finetune VLM + VLM selection & 57.52{\tiny\,$\pm$\,\raisebox{0.45ex}{\rule{2.10em}{0.28pt}}} & 47.45{\tiny\,$\pm$\,\raisebox{0.45ex}{\rule{2.10em}{0.28pt}}} & 34.76{\tiny\,$\pm$\,\raisebox{0.45ex}{\rule{2.10em}{0.28pt}}} \\
 &  & Naive distill + VLM selection & \thirdcell{67.69{\tiny\,$\pm$\,0.16}} & \thirdcell{58.00{\tiny\,$\pm$\,0.41}} & \thirdcell{44.66{\tiny\,$\pm$\,0.97}} \\
 &  & Ours + random selection & \secondcell{69.43{\tiny\,$\pm$\,0.33}} & \secondcell{60.61{\tiny\,$\pm$\,0.30}} & \secondcell{46.06{\tiny\,$\pm$\,0.97}} \\
 &  & Ours + VLM selection & \bestcell{70.71{\tiny\,$\pm$\,0.39}} & \bestcell{62.04{\tiny\,$\pm$\,0.64}} & \bestcell{48.09{\tiny\,$\pm$\,2.20}} \\
\addlinespace[1.0pt]
\midrule
Euclid & 512 & No distil + random selection & 81.19{\tiny\,$\pm$\,1.57} & 71.38{\tiny\,$\pm$\,2.75} & 58.39{\tiny\,$\pm$\,4.73} \\
 &  & Finetune VLM + VLM selection & 69.21{\tiny\,$\pm$\,\raisebox{0.45ex}{\rule{2.10em}{0.28pt}}} & 54.27{\tiny\,$\pm$\,\raisebox{0.45ex}{\rule{2.10em}{0.28pt}}} & 36.83{\tiny\,$\pm$\,\raisebox{0.45ex}{\rule{2.10em}{0.28pt}}} \\
 &  & Naive distill + VLM selection & \secondcell{83.76{\tiny\,$\pm$\,0.52}} & \thirdcell{74.78{\tiny\,$\pm$\,1.42}} & \secondcell{67.58{\tiny\,$\pm$\,2.83}} \\
 &  & Ours + random selection & \thirdcell{82.78{\tiny\,$\pm$\,1.48}} & \secondcell{74.80{\tiny\,$\pm$\,2.26}} & \thirdcell{61.24{\tiny\,$\pm$\,3.88}} \\
 &  & Ours + VLM selection & \bestcell{85.48{\tiny\,$\pm$\,1.01}} & \bestcell{77.68{\tiny\,$\pm$\,1.02}} & \bestcell{69.42{\tiny\,$\pm$\,5.10}} \\
\addlinespace[1.0pt]
 & 2048 & No distil + random selection & 88.73{\tiny\,$\pm$\,0.64} & 81.83{\tiny\,$\pm$\,0.95} & 76.30{\tiny\,$\pm$\,2.59} \\
 &  & Finetune VLM + VLM selection & 72.63{\tiny\,$\pm$\,\raisebox{0.45ex}{\rule{2.10em}{0.28pt}}} & 55.62{\tiny\,$\pm$\,\raisebox{0.45ex}{\rule{2.10em}{0.28pt}}} & 44.63{\tiny\,$\pm$\,\raisebox{0.45ex}{\rule{2.10em}{0.28pt}}} \\
 &  & Naive distill + VLM selection & \thirdcell{88.90{\tiny\,$\pm$\,0.65}} & \thirdcell{82.57{\tiny\,$\pm$\,0.48}} & \secondcell{79.07{\tiny\,$\pm$\,2.42}} \\
 &  & Ours + random selection & \secondcell{89.59{\tiny\,$\pm$\,0.59}} & \secondcell{83.26{\tiny\,$\pm$\,0.39}} & \thirdcell{78.75{\tiny\,$\pm$\,4.00}} \\
 &  & Ours + VLM selection & \bestcell{90.23{\tiny\,$\pm$\,0.24}} & \bestcell{84.72{\tiny\,$\pm$\,0.28}} & \bestcell{80.15{\tiny\,$\pm$\,3.22}} \\
\addlinespace[1.0pt]
 & 8192 & No distil + random selection & 91.30{\tiny\,$\pm$\,0.64} & 85.30{\tiny\,$\pm$\,1.43} & 84.04{\tiny\,$\pm$\,1.16} \\
 &  & Finetune VLM + VLM selection & 75.21{\tiny\,$\pm$\,\raisebox{0.45ex}{\rule{2.10em}{0.28pt}}} & 61.93{\tiny\,$\pm$\,\raisebox{0.45ex}{\rule{2.10em}{0.28pt}}} & 56.28{\tiny\,$\pm$\,\raisebox{0.45ex}{\rule{2.10em}{0.28pt}}} \\
 &  & Naive distill + VLM selection & \secondcell{92.72{\tiny\,$\pm$\,0.58}} & \secondcell{88.04{\tiny\,$\pm$\,0.96}} & \secondcell{87.23{\tiny\,$\pm$\,0.91}} \\
 &  & Ours + random selection & \thirdcell{92.20{\tiny\,$\pm$\,0.45}} & \thirdcell{86.93{\tiny\,$\pm$\,0.92}} & \thirdcell{85.79{\tiny\,$\pm$\,0.80}} \\
 &  & Ours + VLM selection & \bestcell{93.06{\tiny\,$\pm$\,0.26}} & \bestcell{89.37{\tiny\,$\pm$\,0.69}} & \bestcell{87.44{\tiny\,$\pm$\,0.71}} \\
\addlinespace[1.0pt]
\bottomrule
\end{tabular}%
}
\end{table}

We train and evaluate the proposed framework on both survey datasets under
multiple human-label budgets. For each dataset, all galaxies outside the
held-out diagnostic test set are treated as the unlabeled training pool, from
which the labeled subset is selected according to the specified budget. We
compare our method with random-selection baselines on the same test set.
Training and evaluation details are provided in
Appendix~\ref{app:training-evaluation}.
\subsection{Results}
\Cref{tab:main-results} reports the main accuracy results across both survey
domains and all human-label budgets. Our full method, which combines
reachability-aware VLM distillation, residual side tuning, and tree-aware
sample selection, achieves the best performance in every
survey--budget--metric setting. The improvements are particularly strong for
balanced and weak-branch accuracy, indicating that the gains are concentrated
on underrepresented morphology categories and difficult downstream questions
rather than only on common root-level decisions.

The baselines separate the contributions of the individual components.
Finetuning the VLM directly performs substantially worse than the
Zoobot-based methods, confirming that the general-purpose VLM is more effective
as a teacher than as the final morphology classifier. Naive distillation,
which omits both reachability weighting and side tuning, usually improves over
training without a teacher, showing that the VLM already provides useful weak
supervision. However, our distillation procedure with random selection is
consistently stronger, demonstrating the benefit of preserving Zoobot's
pretrained representation through side tuning and respecting the conditional
question tree through reachability-aware supervision.

Finally, replacing random selection with VLM-guided tree-aware selection yields
the strongest results at every budget. This shows that the teacher contributes
in two complementary ways: its soft predictions improve representation
learning during distillation, while its uncertainty and tree-probability
features identify more informative galaxies for human annotation.

\subsection{Ablations}
\label{sec:ablations}
\begin{table}[t]
\centering\scriptsize
\caption{Distillation ablation. Negative values indicate performance decreases from Our distillation method.}
\label{tab:distillation-ablation}
\begin{tabular}{l r c c c}
\toprule
Ablation & Budget & $\Delta$ Overall (\%) & $\Delta$ Balanced (\%) & $\Delta$ Weak (\%) \\
\midrule
w/o reachability & 512 & -1.76 & -1.74 & -3.64 \\
 & 2048 & -0.80 & -1.32 & -1.96 \\
 & 8192 & -0.56 & -0.66 & -2.21 \\
\addlinespace[2.0pt]
w/o side-tuning & 512 & -1.03 & -1.65 & -0.49 \\
 & 2048 & -1.84 & -2.04 & -1.24 \\
 & 8192 & -2.04 & -2.06 & -1.26 \\
\addlinespace[2.0pt]
Post-finetune VLM teacher & 512 &  -0.02&  -0.47&  -0.97 \\
 & 2048 & -0.46 & -0.70 & -0.86 \\
 & 8192 &  -0.15&  0.13&  0.66 \\
\addlinespace[2.0pt]
\bottomrule
\end{tabular}%
\end{table}

\begin{figure}[t]
  \centering
  \includegraphics[width=\columnwidth]{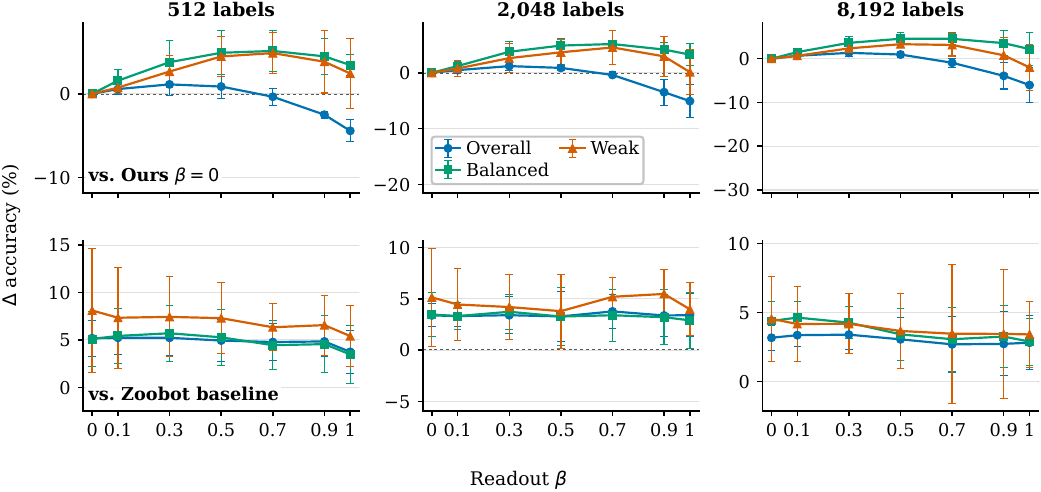}
  \caption{Prior correction ablation. The top row compares our method at each $\beta$ with our method at $\beta=0$; the bottom row compares our method with the baseline at the same $\beta$.}
  \label{fig:beta-ablation}
\end{figure}

\begin{table}[t]
\centering\footnotesize
\caption{Selection-rule ablation. Negative values indicate performance decreases from Our selection method.}
\label{tab:selection-ablation}
\begin{tabular}{l r c c c}
\toprule
Selection variant & Budget & $\Delta$ Overall (\%) & $\Delta$ Balanced (\%) & $\Delta$ Weak (\%) \\
\midrule
Diversity-only & 512 & -0.78 & -0.76 & -2.36 \\
 & 2048 & -0.38 & -0.60 & -1.00 \\
 & 8192 & -0.32 & -0.36 & -0.53 \\
\addlinespace[2.0pt]
Uncertainty-only & 512 & +0.81 & -0.82 & +2.55 \\
 & 2048 & -1.86 & -2.92 & -2.38 \\
 & 8192 & -0.39 & -0.90 & -0.45 \\
\addlinespace[2.0pt]
Zoobot feature & 512 & -2.01 & -1.80 & -3.79 \\
 & 2048 & -0.82 & -0.90 & -0.84 \\
 & 8192 & -1.04 & -0.76 & -1.36 \\
\addlinespace[2.0pt]
\bottomrule
\end{tabular}%
\end{table}

\paragraph{Distillation components.}
\Cref{tab:distillation-ablation} shows that removing either reachability or side tuning will hurt performance.
We further test a co-evolution variant in which the selected human labels are
first used to finetune the VLM teacher, followed by a second round of
teacher-to-student distillation. This post-finetuning teacher provides no
consistent improvement and is slightly worse in most settings. Once the same
human annotations are already used to finetune the student, adapting the
teacher to those labels does not introduce substantial new information into
the teacher--student system. The additional co-evolution stage is therefore
unnecessary in our setting.

\paragraph{Sample-selection rule.}
\Cref{tab:selection-ablation} compares the full uncertainty-and-diversity
criterion with its individual components and alternative features.
Either diversity or uncertainty only selection is worth than them combined. Using pretrained Zoobot image features for diversity selection also performs
worse than our tree-probability representation at every budget. Generic image
features encode many aspects of the scene that are not directly relevant to
the target morphology questions. In contrast, the proposed feature is defined directly over
the predicted question tree and therefore emphasizes variation in the
morphological attributes that the downstream task is designed to classify.

\paragraph{Test-time prior correction.}
\Cref{fig:beta-ablation} studies the correction strength $\beta$.
Mild correction generally improves all accuracy. Moreover, across all correction strength, our method constantly better than baseline.

\section{Conclusion}

We show that general-purpose VLMs contain visual-semantic priors that are useful for galaxy morphology classification and complementary to domain-specific morphology foundation models. Rather than using VLM predictions as final morphology labels, we use Qwen as a weak supervisor for a pretrained Zoobot student. Across two representative telescope domains, this pipeline consistently improves accuracy over baselines across multiple human-label budgets. The proposed VLM-teacher framework therefore is ready to serve as a practical label-efficient component in galaxy morphology analysis pipelines for forthcoming large-scale surveys, including the flagship ground-based LSST and the space-based Roman mission.

\section{Limitations and Ethical Considerations}

This study uses only publicly released astronomical images, survey metadata,
and aggregated Galaxy Zoo vote fractions. The original morphology annotations
were collected through the Zooniverse citizen-science platform, where
volunteers independently answered predefined questions about galaxy images.
We do not access participant identities, individual response histories, or
other personal information, and the study involves no direct interaction with
human participants.

The main limitation concerns the representativeness of the selected survey
domains. Cosmic Dawn and Euclid provide representative examples of ground- and
space-based imaging and approximate several characteristics of forthcoming
LSST and Roman observations, respectively. However, they do not fully reproduce
all characteristics of
those surveys. Moreover, our experiments do not guarantee how either the VLM itself or
the proposed teacher--student framework will generalize to future surveys with substantially more fine-grained morphology questions. However, the potential utility of a VLM teacher on a new survey can be assessed through a small-scale zero-shot evaluation. Our results further show that even VLM predictions containing only weak signal above random guessing can provide useful supervision and improve an already strong domain-specific astronomy foundation model.
\section{Generative AI Usage}
The authors used generative AI tools to assist with training-environment setup
and the implementation of selected utility functions for data processing,
model training, evaluation, and visualization. All core methodological
components were designed and implemented by the authors. Generative AI tools
were also used for language editing and to improve manuscript clarity. All
AI-assisted code and text were inspected, tested, verified, and revised by the
authors as needed. These tools were not used to autonomously determine the
research methodology, interpret experimental results, or formulate the
scientific conclusions. The authors take full responsibility for the final content of the code and manuscript.

{
    \small
    \bibliographystyle{ACM-Reference-Format}
    \bibliography{main}
}

\appendix
\section{VLM-Ready Dataset Construction}
\label{app:dataset-build}

We construct the benchmark from the public Galaxy Zoo: Cosmic Dawn and
Euclid Q1 catalogue and image releases. Because the two surveys use different
catalogue fields and morphology schemas, we standardize them into a shared
decision-tree-aware representation for both VLM evaluation and morphology
classifier training.

For each survey, we match catalogue objects to their image cutouts and map the
survey-specific morphology columns to a canonical schema covering the 13
Galaxy Zoo questions used in our experiments. Each task is represented by a
common identifier, natural-language question, legal answer set, and standardized
vote-fraction columns.

For every valid galaxy--question pair, we retain the full vote-fraction vector
\begin{equation}
    \mathbf{y}_{i,q}
    =
    \left(y_{i,q,1},\ldots,y_{i,q,|\mathcal{C}_q|}\right),
\end{equation}
and derive the hard target
\begin{equation}
    \hat{a}_{i,q}
    =
    \arg\max_{c\in\mathcal{C}_q} y_{i,q,c}.
\end{equation}
Missing or invalid questions remain undefined. Retaining the soft fractions
preserves annotation ambiguity and supports both soft-label training and
distribution-sensitive evaluation.

We further encode the conditional Galaxy Zoo protocol as a machine-readable
question tree containing each question, its legal answers, and its parent
conditions. For each galaxy, the tree identifies the reachable questions and
generates the corresponding VLM instances:
\begin{equation}
    \left(
    \text{image},\ \text{question},\ \text{legal options},\
    \text{path context}
    \right).
\end{equation}
The same standardized galaxy-level representation therefore supports
tree-aware VLM prompting and evaluation as well as multi-head morphology
classifier training.
\begin{table}[t]
\centering
\caption{Comparison of the telescope of our benchmark with
forthcoming LSST and Roman observations.}
\label{tab:survey-comparison}
\scriptsize
\begin{tabular}{lcccc}
\toprule
Parameter
& Cosmic Dawn
& Euclid Q1
& LSST
& Roman \\
\midrule
Facility
& Subaru/HSC
& Euclid/VIS
& Rubin/LSST
& Roman/WFI \\
Regime
& Ground optical
& Space visible
& Ground optical
& Space near-IR \\
Bands
& $g,r,i,z,y$
& $I_{\rm E}$
& $u,g,r,i,z,y$
& $R,Z,Y,J,H,F$ \\
PSF 
& $\sim0.6$--$0.9''$
& $\sim0.16''$
& $\sim0.7''$
& $\sim0.1$--$0.2''$ \\
Pixel scale
& $0.168''$
& $0.1''$
& $0.2''$
& $0.11''$ \\
Depth
& $i\sim27.6$
& $I_{\rm E}\sim25.6$
& $r\sim27.5$--$27.8$
& $H\sim26$--$27$ \\
Area
& $\sim6~{\rm deg}^{2}$
& $63.1~{\rm deg}^{2}$
& $\sim18{,}000~{\rm deg}^{2}$
& $\sim2{,}000~{\rm deg}^{2}$ \\
\bottomrule
\end{tabular}
\end{table}
\section{Tree-balanced evaluation-set construction}
\label{app:tree-balanced-test}

 For each galaxy, the valid hard reference
answers define a deterministic path through the shared morphology tree. Its
strata are all pairs $(q,c)$ on that path.
Given a survey table $D$, morphology tree $T$, maximum row count $M$, target
per answer $K$, let $A_a$ be the galaxies whose paths contain
stratum $a$. We set its attainable target to
\begin{equation}
    k_a=\min(K,|A_a|).
\end{equation}
Starting from an empty selected set, the sampler visits strata from rare to
common and repeatedly adds the candidate with the largest coverage gain over
all currently unmet strata. Sampling stops when every attainable target is
met, no candidates remain, or the set reaches $M$ galaxies. 

This procedure yields 1,066 Cosmic Dawn galaxies covering 44 reachable strata
and 1,168 Euclid galaxies covering 43 reachable strata, with no unmet
attainable stratum in either survey.

\section{Evaluation protocol and metrics}
\label{app:vqa-eval-metrics}

We evaluate each model question-by-question on the reference-reachable path. The valid options are mapped to single-letter
choices, and we apply a softmax to the model logits of those letter tokens at
the answer position. This produces an option distribution $P_{iq}$ and a hard
prediction $\hat y_{iq}$. We compare the prediction with the reference hard
target $y_{iq}$ and, for distribution metrics, compare $P_{iq}$ with the
source fraction vector $F_{iq}$.

Let $E_q$ be the scored examples for question $q$, $C_q^{\mathrm{obs}}$ its
observed reference options, and $N_{qc}$ the number of examples with reference
answer $c$. Overall accuracy pools all scored question instances:
\begin{equation}
    \mathrm{Acc}_{\mathrm{overall}}
    =
    \frac{\sum_q\sum_{i\in E_q}
    \mathbb{1}[\hat y_{iq}=y_{iq}]}
    {\sum_q |E_q|}.
\end{equation}
Balanced accuracy gives every observed answer equal weight within each
question, then macro-averages across the evaluated question set $\mathcal Q'$:
\begin{equation}
    \mathrm{ACC}_{\mathrm{balanced}}
    =
    \frac{1}{|\mathcal Q'|}
    \sum_{q\in\mathcal Q'}
    \frac{1}{|C_q^{\mathrm{obs}}|}
    \sum_{c\in C_q^{\mathrm{obs}}}
    \frac{1}{N_{qc}}
    \sum_{\substack{i\in E_q\\y_{iq}=c}}
    \mathbb{1}[\hat y_{iq}=c].
\end{equation}
Weak-branch accuracy macro-averages question accuracy over artifact type,
bulge size, spiral winding, and spiral arm count:
\begin{equation}
    \mathrm{ACC}_{\mathrm{weak}}
    =
    \frac{1}{|\mathcal W|}
    \sum_{q\in\mathcal W}\mathrm{Acc}_q.
\end{equation}
Finally, Brier score, soft cross-entropy, and Jensen--Shannon divergence measure
agreement with the full source fraction vectors:
{\small
\begin{align}
    N_{\mathrm{dist}}
    &= \sum_q |E_q|,\\
    \mathrm{Brier}
    &=\frac{1}{N_{\mathrm{dist}}}\sum_q\sum_{i\in E_q}
      \sum_{c\in C_q}(P_{iqc}-F_{iqc})^2,\\
    \mathrm{SoftCE}
    &=-\frac{1}{N_{\mathrm{dist}}}\sum_q\sum_{i\in E_q}
      \sum_{c\in C_q}F_{iqc}\log\!\left(\max(P_{iqc},10^{-6})\right),\\
    \mathrm{JS}
    &=\frac{1}{N_{\mathrm{dist}}}\sum_q\sum_{i\in E_q}
      \frac{\mathrm{KL}(F_{iq}\|M_{iq})+\mathrm{KL}(P_{iq}\|M_{iq})}{2},
    \qquad
    M_{iq}=\frac{F_{iq}+P_{iq}}{2}.
\end{align}
}

\section{Prompt-Engineering Ablation}
\label{app:prompt-engineering}

\subsection{Prompt Designs}
\label{app:prompt-designs}

The templates below show the structure of each selectable inference prompt.
Exact question text, option labels, morphology cards, physical
grounding, decision-tree history, and survey-specific context are available in our open codebase.

\newcommand{\promptslot}[1]{\textnormal{\textless#1\textgreater}}
\newtcolorbox{promptbox}[1]{
    enhanced,
    breakable,
    colback=gray!3,
    colframe=gray!50!black,
    boxrule=0.6pt,
    arc=6pt,
    fonttitle=\bfseries,
    title={#1},
    left=8pt,right=8pt,top=6pt,bottom=6pt
}

\subsubsection{V0: Basic (\textnormal{\texttt{v0\_baseline}})}
\begin{promptbox}{Prompt Design for Morphology VQA: V0 Basic}
\promptslot{SYSTEM PROMPT} You are a galaxy morphology expert.\\
Choose the best Galaxy Zoo answer using only visible morphology in the image.\\[0.4ex]
\promptslot{USER PROMPT} Question: \{Galaxy Zoo question\}\\
Options:\\
A. \{answer label 1\}\\
B. \{answer label 2\}\\
C. \{answer label 3\}\\
\ldots\\[0.4ex]
Return only the single option letter, with no explanation.
\end{promptbox}

\subsubsection{V1: Phenomenology (\textnormal{\texttt{v1\_pheno}})}
\begin{promptbox}{Prompt Design for Morphology VQA: V1 Phenomenology}
\promptslot{SYSTEM PROMPT} You are a galaxy morphology expert.\\
Choose the best Galaxy Zoo answer using only visible morphology in the image.\\[0.4ex]
\promptslot{VISIBLE-MORPHOLOGY RULES} \{central-object, visible-evidence, and resolution rules\}\\[0.4ex]
\promptslot{USER PROMPT} Question: \{Galaxy Zoo question\}\\
Options:\\
A. \{answer label 1\} --- \{visible-morphology card for answer 1\}\\
B. \{answer label 2\} --- \{visible-morphology card for answer 2\}\\
C. \{answer label 3\} --- \{visible-morphology card for answer 3\}\\
\ldots\\[0.4ex]
Return only the single option letter, with no explanation.
\end{promptbox}

\subsubsection{V2: Phenomenology + Physics (\textnormal{\texttt{v2\_pheno\_phys}})}
\begin{promptbox}{Prompt Design for Morphology VQA: V2 Phenomenology + Physics}
\promptslot{SYSTEM PROMPT} You are a galaxy morphology expert.\\
Choose the best Galaxy Zoo answer using only visible morphology in the image.\\[0.4ex]
\promptslot{VISIBLE-MORPHOLOGY RULES} \{central-object, visible-evidence, and resolution rules\}\\[0.4ex]
\promptslot{USER PROMPT} Question: \{Galaxy Zoo question\}\\
Options:\\
A. \{answer label 1\} --- \{visible-morphology card\}; Physical grounding: \{physics card\}\\
B. \{answer label 2\} --- \{visible-morphology card\}; Physical grounding: \{physics card\}\\
C. \{answer label 3\} --- \{visible-morphology card\}; Physical grounding: \{physics card\}\\
\ldots\\[0.4ex]
Return only the single option letter, with no explanation.
\end{promptbox}

\subsubsection{V3: Evidence First (\textnormal{\texttt{v3\_evidence}})}
\begin{promptbox}{Prompt Design for Morphology VQA: V3 Evidence First}
\promptslot{SYSTEM PROMPT}  You are a galaxy morphology expert.\\
Choose the best Galaxy Zoo answer using only visible morphology in the image.\\[0.4ex]
\promptslot{USER PROMPT: EVIDENCE PASS} Before choosing, write concise visual evidence
for each option letter. Use \emph{none} when an option has no visible support.
Do not provide a final answer yet.\\[0.6ex]
\promptslot{MODEL EVIDENCE} \{generated evidence for options A, B, C, \ldots\}\\[0.6ex]
\promptslot{USER PROMPT: SCORING PASS} Considering the evidence, return only the
single option letter.\\
\end{promptbox}

\subsubsection{V4:  Phenomenology + Physics  + Survey Context (\textnormal{\texttt{v4\_context}})}
\begin{promptbox}{Prompt Design for Morphology VQA: V4 Survey Context}
\promptslot{SYSTEM PROMPT} You are a galaxy morphology expert.\\
Choose the best Galaxy Zoo answer using only visible morphology in the image.\\[0.4ex]
\promptslot{VISIBLE-MORPHOLOGY RULES} \{central-object, visible-evidence, and resolution rules\}\\[0.4ex]
\promptslot{SURVEY CONTEXT} \{Euclid or Cosmic Dawn imaging characteristics and artifact context\}\\[0.4ex]

\promptslot{USER PROMPT} Question: \{Galaxy Zoo question\}\\
Options:\\
A. \{answer label 1\} --- \{visible-morphology card\}; Physical grounding: \{physics card\}\\
B. \{answer label 2\} --- \{visible-morphology card\}; Physical grounding: \{physics card\}\\
C. \{answer label 3\} --- \{visible-morphology card\}; Physical grounding: \{physics card\}\\
\ldots\\[0.4ex]
Return only the single option letter, with no explanation.
\end{promptbox}

\section{Training Details}
\label{app:training-evaluation}

We use Qwen3.5-9B as the VLM teacher and Zoobot as the morphology student.
For main experiments, VLM teacher are frozen. For ablation of finetuning VLM and post-finetune VLM teacher, Qwen is adapted with LoRA while all pretrained Qwen parameters remain
frozen. LoRA modules are inserted into the attention and MLP projections. We use rank
$r=16$, scaling $\alpha=32$, and dropout $0.05$, resulting in 29.1M trainable
parameters out of 9.44B total parameters (0.308\%). The adapters are trained
for one epoch with effective batch size 16 using AdamW, a learning rate of
$2\times10^{-4}$, zero weight decay, gradient clipping at norm 1.0, and a
3\% linear warm-up followed by a constant learning rate.

For the student, we initialize the Cosmic Dawn experiments from Zoobot v1 with
an EfficientNet-B0 encoder and the Euclid experiments from Zoobot v2 with a
ConvNeXt-Nano encoder. This choice avoids potential pretraining leakage for
Cosmic Dawn, which is represented in the Zoobot v2 pretraining data, while
Zoobot v1 predates the Cosmic Dawn release. Images are resized to $224\times224$ and normalized. During VLM
distillation, the Zoobot encoder is frozen and only the residual side adapter
and question-specific heads are optimized. We train for three epochs with
batch size 512 using AdamW at a learning rate of $10^{-4}$. The side adapter uses bottleneck width 128, residual scale
1.0, and a zero-initialized output projection.

For human-label finetuning, we unfreeze the full student and optimize it for
200 updates using vote-fraction cross-entropy. We evaluate label budgets
$B\in\{512,2048,8192\}$ and report results over three random seeds for each
setting. All experiments are run on a single NVIDIA H100 GPU.

\section{Additional Results}
\Cref{tab:distribution-results} reports distribution-sensitive comparison
metrics across surveys and label budgets. \Cref{tab:zero-shot-per-question}
summarizes zero-shot accuracy for each morphology question. 
\Cref{tab:zero-shot-per-class,tab:zero-shot-per-class-continued} provide the
corresponding option-level zero-shot accuracies for Questions Q01--Q13.
\begin{table}[t]
\centering\scriptsize
\setlength{\tabcolsep}{1.8pt}
\caption{Main comparison distribution metrics. Finetune VLM is a single reported value. Colors mark rank by the mean within each survey-budget-metric block.}
\label{tab:distribution-results}
\resizebox{\columnwidth}{!}{%
\begin{tabular}{l r l c c c}
\toprule
Survey & Budget & Method & Brier \(\downarrow\) & Soft CE \(\downarrow\) & JS \(\downarrow\) \\
\midrule
Cosmic Dawn & 512 & No distil + random selection & \thirdcell{0.2286{\tiny\,$\pm$\,0.0119}} & \secondcell{1.0274{\tiny\,$\pm$\,0.0294}} & \thirdcell{0.1103{\tiny\,$\pm$\,0.0042}} \\
 &  & Finetune VLM + VLM selection & 0.3157{\tiny\,$\pm$\,\raisebox{0.45ex}{\rule{3.20em}{0.28pt}}} & 1.3170{\tiny\,$\pm$\,\raisebox{0.45ex}{\rule{3.20em}{0.28pt}}} & 0.1418{\tiny\,$\pm$\,\raisebox{0.45ex}{\rule{3.20em}{0.28pt}}} \\
 &  & Naive distill + VLM selection & \bestcell{0.2084{\tiny\,$\pm$\,0.0031}} & \bestcell{0.9909{\tiny\,$\pm$\,0.0159}} & \secondcell{0.0989{\tiny\,$\pm$\,0.0009}} \\
 &  & Ours + random selection & 0.2712{\tiny\,$\pm$\,0.0124} & 1.2289{\tiny\,$\pm$\,0.0717} & 0.1154{\tiny\,$\pm$\,0.0039} \\
 &  & Ours + VLM selection & \secondcell{0.2204{\tiny\,$\pm$\,0.0073}} & \thirdcell{1.0591{\tiny\,$\pm$\,0.0288}} & \bestcell{0.0966{\tiny\,$\pm$\,0.0023}} \\
\addlinespace[1.0pt]
 & 2048 & No distil + random selection & \thirdcell{0.1852{\tiny\,$\pm$\,0.0054}} & \thirdcell{0.9380{\tiny\,$\pm$\,0.0111}} & 0.0942{\tiny\,$\pm$\,0.0019} \\
 &  & Finetune VLM + VLM selection & 0.3514{\tiny\,$\pm$\,\raisebox{0.45ex}{\rule{3.20em}{0.28pt}}} & 1.3613{\tiny\,$\pm$\,\raisebox{0.45ex}{\rule{3.20em}{0.28pt}}} & 0.1501{\tiny\,$\pm$\,\raisebox{0.45ex}{\rule{3.20em}{0.28pt}}} \\
 &  & Naive distill + VLM selection & \secondcell{0.1836{\tiny\,$\pm$\,0.0036}} & \secondcell{0.9361{\tiny\,$\pm$\,0.0041}} & \thirdcell{0.0902{\tiny\,$\pm$\,0.0013}} \\
 &  & Ours + random selection & 0.1903{\tiny\,$\pm$\,0.0040} & 0.9652{\tiny\,$\pm$\,0.0121} & \secondcell{0.0877{\tiny\,$\pm$\,0.0012}} \\
 &  & Ours + VLM selection & \bestcell{0.1723{\tiny\,$\pm$\,0.0008}} & \bestcell{0.9209{\tiny\,$\pm$\,0.0023}} & \bestcell{0.0808{\tiny\,$\pm$\,0.0006}} \\
\addlinespace[1.0pt]
 & 8192 & No distil + random selection & \thirdcell{0.1784{\tiny\,$\pm$\,0.0058}} & 0.9272{\tiny\,$\pm$\,0.0123} & 0.0910{\tiny\,$\pm$\,0.0022} \\
 &  & Finetune VLM + VLM selection & 0.3271{\tiny\,$\pm$\,\raisebox{0.45ex}{\rule{3.20em}{0.28pt}}} & 1.3027{\tiny\,$\pm$\,\raisebox{0.45ex}{\rule{3.20em}{0.28pt}}} & 0.1372{\tiny\,$\pm$\,\raisebox{0.45ex}{\rule{3.20em}{0.28pt}}} \\
 &  & Naive distill + VLM selection & 0.1784{\tiny\,$\pm$\,0.0024} & \thirdcell{0.9271{\tiny\,$\pm$\,0.0049}} & \thirdcell{0.0882{\tiny\,$\pm$\,0.0008}} \\
 &  & Ours + random selection & \secondcell{0.1708{\tiny\,$\pm$\,0.0030}} & \secondcell{0.9198{\tiny\,$\pm$\,0.0079}} & \secondcell{0.0820{\tiny\,$\pm$\,0.0009}} \\
 &  & Ours + VLM selection & \bestcell{0.1625{\tiny\,$\pm$\,0.0015}} & \bestcell{0.8991{\tiny\,$\pm$\,0.0040}} & \bestcell{0.0788{\tiny\,$\pm$\,0.0007}} \\
\addlinespace[1.0pt]
\midrule
Euclid & 512 & No distil + random selection & 0.0413{\tiny\,$\pm$\,0.0059} & 0.8110{\tiny\,$\pm$\,0.0142} & 0.0195{\tiny\,$\pm$\,0.0036} \\
 &  & Finetune VLM + VLM selection & 0.1695{\tiny\,$\pm$\,\raisebox{0.45ex}{\rule{3.20em}{0.28pt}}} & 1.2363{\tiny\,$\pm$\,\raisebox{0.45ex}{\rule{3.20em}{0.28pt}}} & 0.0821{\tiny\,$\pm$\,\raisebox{0.45ex}{\rule{3.20em}{0.28pt}}} \\
 &  & Naive distill + VLM selection & \secondcell{0.0286{\tiny\,$\pm$\,0.0034}} & \secondcell{0.7833{\tiny\,$\pm$\,0.0059}} & \secondcell{0.0130{\tiny\,$\pm$\,0.0013}} \\
 &  & Ours + random selection & \thirdcell{0.0359{\tiny\,$\pm$\,0.0063}} & \thirdcell{0.7999{\tiny\,$\pm$\,0.0152}} & \thirdcell{0.0168{\tiny\,$\pm$\,0.0036}} \\
 &  & Ours + VLM selection & \bestcell{0.0270{\tiny\,$\pm$\,0.0038}} & \bestcell{0.7813{\tiny\,$\pm$\,0.0073}} & \bestcell{0.0121{\tiny\,$\pm$\,0.0016}} \\
\addlinespace[1.0pt]
 & 2048 & No distil + random selection & 0.0179{\tiny\,$\pm$\,0.0012} & 0.7650{\tiny\,$\pm$\,0.0018} & 0.0087{\tiny\,$\pm$\,0.0005} \\
 &  & Finetune VLM + VLM selection & 0.1978{\tiny\,$\pm$\,\raisebox{0.45ex}{\rule{3.20em}{0.28pt}}} & 1.4885{\tiny\,$\pm$\,\raisebox{0.45ex}{\rule{3.20em}{0.28pt}}} & 0.0964{\tiny\,$\pm$\,\raisebox{0.45ex}{\rule{3.20em}{0.28pt}}} \\
 &  & Naive distill + VLM selection & \secondcell{0.0138{\tiny\,$\pm$\,0.0011}} & \secondcell{0.7574{\tiny\,$\pm$\,0.0033}} & \secondcell{0.0065{\tiny\,$\pm$\,0.0005}} \\
 &  & Ours + random selection & \thirdcell{0.0159{\tiny\,$\pm$\,0.0018}} & \thirdcell{0.7610{\tiny\,$\pm$\,0.0030}} & \thirdcell{0.0077{\tiny\,$\pm$\,0.0008}} \\
 &  & Ours + VLM selection & \bestcell{0.0132{\tiny\,$\pm$\,0.0010}} & \bestcell{0.7555{\tiny\,$\pm$\,0.0014}} & \bestcell{0.0061{\tiny\,$\pm$\,0.0003}} \\
\addlinespace[1.0pt]
 & 8192 & No distil + random selection & 0.0109{\tiny\,$\pm$\,0.0002} & 0.7523{\tiny\,$\pm$\,0.0005} & 0.0056{\tiny\,$\pm$\,0.0001} \\
 &  & Finetune VLM + VLM selection & 0.1507{\tiny\,$\pm$\,\raisebox{0.45ex}{\rule{3.20em}{0.28pt}}} & 1.2274{\tiny\,$\pm$\,\raisebox{0.45ex}{\rule{3.20em}{0.28pt}}} & 0.0744{\tiny\,$\pm$\,\raisebox{0.45ex}{\rule{3.20em}{0.28pt}}} \\
 &  & Naive distill + VLM selection & \secondcell{0.0073{\tiny\,$\pm$\,0.0006}} & \bestcell{0.7452{\tiny\,$\pm$\,0.0012}} & \secondcell{0.0037{\tiny\,$\pm$\,0.0003}} \\
 &  & Ours + random selection & \thirdcell{0.0090{\tiny\,$\pm$\,0.0002}} & \thirdcell{0.7487{\tiny\,$\pm$\,0.0004}} & \thirdcell{0.0046{\tiny\,$\pm$\,0.0001}} \\
 &  & Ours + VLM selection & \bestcell{0.0073{\tiny\,$\pm$\,0.0006}} & \secondcell{0.7457{\tiny\,$\pm$\,0.0011}} & \bestcell{0.0036{\tiny\,$\pm$\,0.0003}} \\
\addlinespace[1.0pt]
\bottomrule
\end{tabular}%
}
\end{table}

\begin{table}[t]
  \centering
  \caption{Zero-shot per-question accuracy (\%).}
  \label{tab:zero-shot-per-question}
  \footnotesize
  \setlength{\tabcolsep}{2.2pt}
  \renewcommand{\arraystretch}{1.05}
  \begin{tabular}{lccc@{\hspace{5pt}}ccc}
    \toprule
    & \multicolumn{3}{c}{Cosmic Dawn} & \multicolumn{3}{c}{Euclid} \\
    \cmidrule(lr){2-4}\cmidrule(l){5-7}
    Question & Qwen & Random & Human & Qwen & Random & Human \\
    \midrule
    Q01 & 44.84 & 33.00 & 68.00 & 64.47 & 33.00 & 76.00 \\
    Q02 & 58.85 & 33.00 & 83.00 & 78.13 & 33.00 & 78.00 \\
    Q03 & 82.81 & 50.00 & 92.00 & 90.00 & 50.00 & 95.00 \\
    Q04 & 36.65 & 33.00 & 75.00 & 33.33 & 33.00 & 54.00 \\
    Q05 & 32.29 & 50.00 & 84.00 & 54.91 & 50.00 & 86.00 \\
    Q06 & 67.71 & 33.00 & 70.00 & 70.54 & 33.00 & 63.00 \\
    Q07 & 30.07 & 20.00 & 59.00 & 48.88 & 20.00 & 54.00 \\
    Q08 & 24.86 & 33.00 & 67.00 & 24.22 & 33.00 & 62.00 \\
    Q09 & 18.03 & 17.00 & 60.00 & 16.67 & 17.00 & 44.00 \\
    Q10 & 65.38 & 25.00 & 69.00 & 71.15 & 25.00 & 58.00 \\
    Q11 & 70.23 & 50.00 & 79.00 & 61.38 & 50.00 & 67.00 \\
    Q12 & 44.44 & 33.00 & 78.00 & 40.77 & 33.00 & 74.00 \\
    Q13 & 43.40 & 14.00 & 67.00 & 29.33 & 14.00 & 57.00 \\
    \bottomrule
  \end{tabular}
\end{table}

\begin{table}[H]
  \centering
  \caption{Zero-shot per-class accuracy (\%; Questions Q01--Q07). ``--'' denotes an unavailable evaluation.}
  \label{tab:zero-shot-per-class}
  \scriptsize
  \setlength{\tabcolsep}{2.4pt}
  \renewcommand{\arraystretch}{0.98}
  \begin{tabular}{llccc@{\hspace{4pt}}ccc}
    \toprule
    & & \multicolumn{3}{c}{Cosmic Dawn} & \multicolumn{3}{c}{Euclid} \\
    \cmidrule(lr){3-5}\cmidrule(l){6-8}
    Question & Answer & Qwen & Random & Human & Qwen & Random & Human \\
    \midrule
    Q01 & smooth     & 63.54 & 33.00 & 68.00 & 52.60 & 33.00 & 69.00 \\
    Q01 & featured   & 53.59 & 33.00 & 64.00 & 95.47 & 33.00 & 75.00 \\
    Q01 & problem    & 5.56  & 33.00 & 76.00 & 12.20 & 33.00 & 83.00 \\
    \midrule
    Q02 & round      & 98.44 & 33.00 & 86.00 & 96.88 & 33.00 & 80.00 \\
    Q02 & in-between & 23.44 & 33.00 & 83.00 & 48.44 & 33.00 & 77.00 \\
    Q02 & cigar      & 54.69 & 33.00 & 80.00 & 89.06 & 33.00 & 76.00 \\
    \midrule
    Q03 & yes        & 46.07 & 50.00 & 86.00 & 91.15 & 50.00 & 93.00 \\
    Q03 & no         & 98.44 & 50.00 & 94.00 & 89.51 & 50.00 & 96.00 \\
    \midrule
    Q04 & rounded    & 78.12 & 33.00 & 83.00 & 98.44 & 33.00 & 67.00 \\
    Q04 & boxy       & 0.00  & 33.00 & 68.00 & 0.00  & 33.00 & 42.00 \\
    Q04 & none       & 31.25 & 33.00 & 74.00 & 1.56  & 33.00 & 53.00 \\
    \midrule
    Q05 & yes        & 16.94 & 50.00 & 83.00 & 48.70 & 50.00 & 90.00 \\
    Q05 & no         & 100.00& 50.00 & 90.00 & 92.19 & 50.00 & 63.00 \\
    \midrule
    Q06 & strong     & 0.00  & 33.00 & 56.00 & 0.00  & 33.00 & 47.00 \\
    Q06 & weak       & 0.00  & 33.00 & 57.00 & 0.00  & 33.00 & 50.00 \\
    Q06 & no         & 100.00& 33.00 & 76.00 & 100.00& 33.00 & 69.00 \\
    \midrule
    Q07 & none       & 63.77 & 20.00 & 57.00 & 46.97 & 20.00 & 39.00 \\
    Q07 & small      & 11.54 & 20.00 & 64.00 & 17.14 & 20.00 & 58.00 \\
    Q07 & moderate   & 42.62 & 20.00 & 56.00 & 76.19 & 20.00 & 62.00 \\
    Q07 & large      & 0.00  & 20.00 & 52.00 & 0.00  & 20.00 & 40.00 \\
    Q07 & dominant   & 37.50 & 20.00 & 66.00 & --    & 20.00 & --    \\
    \bottomrule
  \end{tabular}
\end{table}

\begin{table}[H]
  \centering
  \caption{Zero-shot per-class accuracy (\%; continued; Questions Q08--Q13).}
  \label{tab:zero-shot-per-class-continued}
  \scriptsize
  \setlength{\tabcolsep}{2.4pt}
  \renewcommand{\arraystretch}{0.98}
  \begin{tabular}{llccc@{\hspace{4pt}}ccc}
    \toprule
    & & \multicolumn{3}{c}{Cosmic Dawn} & \multicolumn{3}{c}{Euclid} \\
    \cmidrule(lr){3-5}\cmidrule(l){6-8}
    Question & Answer & Qwen & Random & Human & Qwen & Random & Human \\
    \midrule
    Q08 & tight      & 0.00   & 33.00 & 69.00 & 0.00   & 33.00 & 64.00 \\
    Q08 & medium     & 0.00   & 33.00 & 59.00 & 0.00   & 33.00 & 46.00 \\
    Q08 & loose      & 100.00 & 33.00 & 74.00 & 100.00 & 33.00 & 71.00 \\
    \midrule
    Q09 & 1          & 0.00  & 17.00 & 65.00 & 0.00  & 17.00 & 55.00 \\
    Q09 & 2          & 23.44 & 17.00 & 84.00 & 60.94 & 17.00 & 72.00 \\
    Q09 & 3          & 0.00  & 17.00 & 55.00 & 0.00  & 17.00 & 39.00 \\
    Q09 & 4          & 0.00  & 17.00 & 42.00 & 1.56  & 17.00 & 28.00 \\
    Q09 & $>4$       & 0.00  & 17.00 & 50.00 & 0.00  & 17.00 & 31.00 \\
    Q09 & can't tell & 79.69 & 17.00 & 61.00 & 37.50 & 17.00 & 42.00 \\
    \midrule
    Q10 & none       & 89.10 & 25.00 & 72.00 & 85.82 & 25.00 & 66.00 \\
    Q10 & minor      & 1.01  & 25.00 & 64.00 & 26.56 & 25.00 & 32.00 \\
    Q10 & major      & 8.05  & 25.00 & 54.00 & 32.97 & 25.00 & 41.00 \\
    Q10 & merger     & 37.66 & 25.00 & 66.00 & 53.98 & 25.00 & 47.00 \\
    \midrule
    Q11 & yes        & 37.60 & 50.00 & 71.00 & 68.38 & 50.00 & 68.00 \\
    Q11 & no         & 86.81 & 50.00 & 83.00 & 52.31 & 50.00 & 66.00 \\
    \midrule
    Q12 & star       & 96.88 & 33.00 & 80.00 & 92.19 & 33.00 & 80.00 \\
    Q12 & artifact   & 11.32 & 33.00 & 71.00 & 29.33 & 33.00 & 74.00 \\
    Q12 & zoom       & 46.88 & 33.00 & 86.00 & 26.56 & 33.00 & 71.00 \\
    \midrule
    Q13 & satellite  & 0.00  & 14.00 & 100.00 & 71.88 & 14.00 & 52.00 \\
    Q13 & scattered  & 18.18 & 14.00 & 59.00  & --    & 14.00 & --    \\
    Q13 & diffraction& 0.00  & 14.00 & 63.00  & 0.00  & 14.00 & 33.00 \\
    Q13 & ray        & --    & 14.00 & --     & --    & 14.00 & --    \\
    Q13 & saturation & 0.00  & 14.00 & 54.00  & 0.00  & 14.00 & 33.00 \\
    Q13 & other      & 71.43 & 14.00 & 74.00  & 21.88 & 14.00 & 57.00 \\
    Q13 & ghost      & --    & 14.00 & --     & 1.56  & 14.00 & 66.00 \\
    \bottomrule
  \end{tabular}
\end{table}

\end{document}